\documentclass[a4paper]{article}

\usepackage{INTERSPEECH2018}
\usepackage{times}
\usepackage{latexsym}
\usepackage{tikz-dependency}
\usepackage{amssymb}
\usepackage{amsmath}
\usepackage{url}
\usepackage[framemethod=TikZ]{mdframed}
\usepackage{mathtools}
\usepackage{floatrow}
\usepackage{subcaption}
\usepackage{algorithmicx}
\usepackage[ruled]{algorithm}
\usepackage[noend]{algpseudocode}

\newcommand{\norm}[1]{\left\lVert#1\right\rVert_2}

\DeclarePairedDelimiter\floor{\lfloor}{\rfloor}

\title{Entity-Aware Language Model as an Unsupervised Reranker}
\name{Mohammad Sadegh Rasooli$^1$, Sarangarajan Parthasarathy$^2$}
\address{
  $^1$Department of Computer Science, Columbia University, New York, NY, 10027   \\
  $^2$Artificial Intelligence and Research, Microsoft Corporation, USA }
\email{rasooli@cs.columbia.edu, sarangp@microsoft.com }

\begin{document}
\maketitle

%\begin{abstract}
%Language modeling is a useful step in reranking the n-best output from a speech recognition system. Unfortunately, language %models are not explicitly able to encode information about the world entities. One solution is to provide transcribed n-best %list in which a human labels the best output in the n-best list such that one can train a reranker with global features on %it. In the absence of such data, we propose a method to alleviate the need of transcribed data. Our model based on the %contrastive estimation method ~\cite{smith2005contrastive}, uses artificial n-best lists to train a reranking model. Thus we %are able to use global features as well as features extracted from the knowledge-base. Our experiments on the music domain, %shows an improvement on the word error rate both on the holdout data and blind test data.
%\end{abstract}

\begin{abstract}
%Reranking n-best hypothesis generated by a speech recognizer using language models %which are
%impractical to invoke in the first-pass, 
%has been shown to be effective in improving
%recognition accuracy. 
In language modeling, it is difficult to incorporate entity relationships from
a knowledge-base. One solution is to use a reranker trained with global features, in which global features are derived from n-best lists.
However, training such a reranker requires manually annotated n-best lists, which is expensive to obtain.
We propose a method based on the contrastive estimation method that alleviates the need for such data. Experiments in the music domain demonstrate that global features, as well as features extracted from an external knowledge-base, can be incorporated into our reranker. Our final model, a simple ensemble of a language model and reranker, achieves a 0.44\% absolute word error rate improvement over an LSTM language model on the blind test data. 
\end{abstract}

\section{Introduction}
\begin{figure}[t!]
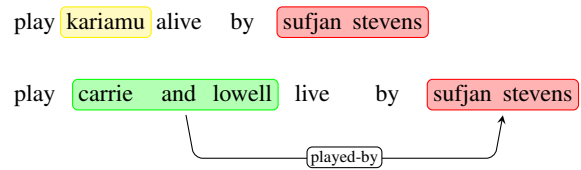

    \centering

\scalebox{1}{
\begin{dependency}[theme = default]
\begin{deptext}
play \& kariamu \& alive  \& by \&  sufjan \& stevens \\ 
\\
\\
   play \&  carrie \& and \& lowell \& live \& by \& sufjan \& stevens \\
\end{deptext}

\wordgroup[group style={fill=yellow!30, draw=yellow}]{1}{2}{2}{singer0}			
\wordgroup[group style={fill=red!30, draw=red}]{1}{5}{6}{singer1}
\wordgroup[group style={fill=green!30, draw=green}]{4}{2}{4}{song}
\wordgroup[group style={fill=red!30, draw=red}]{4}{7}{8}{singer}
\groupedge[edge below]{song}{singer}{played-by}{4ex}

%\groupedge[edge below]{pred}{a0}{ARG0}{4ex}
\end{dependency}
  }
    \caption{An example of two possible outputs by the speech recognition system. The LSTM-based language model predicts the first option while our model prefers the correct output. }
    \label{fig_example1}
  
\end{figure}

Voice assistant systems rely heavily on complex language models. These language models are used as a second-pass reranking step for hypotheses generated by a first-pass speech recognizer~\cite{bengio2003neural,sundermeyer2012lstm,melis2017state,Biadsy2017maxent}. While a significant fraction of queries in voice assistant systems relate to entities in the real world (such as names of places, products, arts, people, etc.), language models do not explicitly model them. The problem of recognizing entities becomes more critical when some entities are not adequately represented in the training data. For example, Figure~\ref{fig_example1} shows two possible outputs from a first-pass speech recognition system. The correct output is about playing a song by a specific singer. The singer and the song have not been seen in the training data together, but by incorporating the cross-entity relationship from a relevant knowledge-base, we can capture the correct output of the speech recognition system.

Unlike previous work~\cite{levit2015token,ahn2016neural} that have considered modeling entities directly in a language model, this paper proposes a dynamic reranking approach without the need for any transcribed training data. Our model makes use of raw text and a knowledge-base consisting of entities in the world. In order to train a reranker, we create artificial n-best lists for each training sentence. This enables us to train a reranking model on the artificial n-best list. We use the contrastive estimation method~\cite{smith2005contrastive}\footnote{This method is different from the noise contrastive estimation method \cite{pmlr-v9-gutmann10a}.} to maximize the likelihood of each sentence in the training data in contrast to artificial sentences in the n-best list. One main strength of our method is that we are not bound to local word-based features anymore, thereby facilitating the possibility of embedding global features such as phrase-based interactions between entities (such as the ``played-by'' relationship in Figure~\ref{fig_example1}). 

This paper considers using language modeling as a global reranking approach in which the reranker makes use of many features including the bidirectional LSTM-based representations of sentences, n-gram language model probability, cross-entity relationships and frequency of entities in the knowledge-base. In other words, we maximize the probability of the whole sentence given its artificial negative examples. By employing this approach, we are able to improve the word error rate of our in-house speech recognition system by $0.44~\%$ absolute difference. This reduction is in particular very interesting to us, because we only target one aspect of the language modeling problem.

In summary, the main contributions of this paper are as following:

\begin{itemize}
    \item Designing a reranker, based on global features, to incorporate entities in the language model.
    \item Proposing an effective approach for generating artificial n-best lists for training a reranker. By applying this idea, we do not need to have transcribed training data.
    \item Introducing features from a knowledge-base and showing their effectiveness in the performance of our speech recognition system.
\end{itemize}

The remainder of this paper is as following: \S\ref{sec_back} briefly describes the background on language modeling. \S\ref{sec_approach} describes our approach and \S\ref{sec_experiments} shows the experimental results. Finally, we conclude in \S\ref{sec_conclusion}.
\section{Background}\label{sec_back}

A recurrent language model uses a recurrent function $RNN$ and derives the intermediate representation $h_i = RNN(h_{i-1}; \theta) \in \mathbb{R}^H$ for every word $x_i$ in sentence $X$. $\theta$ is a set of parameters in the recurrent model.
\begin{equation}\label{eq_lang_model}
	p(X ; \theta)   =  \prod_{i=1}^{n} p(x_i|h_{i-1}; \theta)   = \prod_{i=1}^{n} \sigma_{x_i}(W h_{i-1}+b) 
\end{equation}
In Eq.~\ref{eq_lang_model}, $\sigma$ is the softmax function. $W \in \mathbb{R}^{V \times H}$ and $b\in \mathbb{R}^{V}$ are the output layer and bias term, and $V$ is the vocabulary size. 

Recent studies~\cite{melis2017state,mikolov2011extensions,mikolov2012context} have shown that recurrent neural network (RNN) language models perform significantly better than n-gram language models. Long short-term memories (LSTM)~\cite{hochreiter1997long} are the most popular RNN functions used in the language modeling problem.  Besides all the benefits in using recurrent models, they can only compute sentence probabilities in word space; this is an important problem especially for multi-word entities. 

Representing entities in language models has been considered in previous work. For example, entities are modeled by \cite{levit2015token,levit2015personalization} as additional information in an ngram language model.   Ahn et al. \cite{ahn2016neural} uses a fact-based model to incorporate information available in a knowledge base.  In contrast to their word-based model, our model can capture global information beyond words. Ji et al. \cite{ji2017dynamic} uses a dynamic model to incorporate multiple entities while processing the data. Their model achieves a slight improvement in perplexity, but it is not clear if their model can improve the error rate on big datasets. The recent work by Biadsy et al. \cite{Biadsy2017maxent} shows the effectiveness of log-linear models with global features using transcribed data. We instead use the contrastive estimation method~\cite{smith2005contrastive} to maximize the probability of a correct sentence given its implicit negative examples. This enables us to use global features in our model, as well as word-based features, without using any transcribed data.
\section{Approach}\label{sec_approach}

This section describes our approach based on the contrastive estimation method~\cite{smith2005contrastive}. We list the features and describe the neural network architecture that we use in our model. 

\begin{figure}[t!]
    \centering

\scalebox{1}{
\begin{dependency}[theme = default]
\begin{deptext}
   jim \& dandy's \& near \& me\&  noblesville \&  indiana \\
   ~ \\
   gym \& dandies \& near \& me \& noblesville \&  indiana \\ 
   ~ \\
   jim \& dandys \&  near\&  me \& noblesville \& indiana \\ \\
   jim \&  dandes \&  near\&  me \& noblesville \& indiana \\ \\ 
   jim \& dandies \&  near\&  me \& noblesville \& indiana  \\
\end{deptext}

\wordgroup[group style={fill=green!30, draw=white}]{1}{1}{2}{s0}			

\wordgroup[group style={fill=red!30, draw=white}]{3}{1}{2}{s1}			
\wordgroup[group style={fill=red!30, draw=white}]{5}{2}{2}{s2}			
\wordgroup[group style={fill=red!30, draw=white}]{7}{2}{2}{s3}			
\wordgroup[group style={fill=red!30, draw=white}]{9}{2}{2}{s4}

%\groupedge[edge below]{pred}{a0}{ARG0}{4ex}
\end{dependency}
  }

    \caption{An example of a correct sentence on top and its implicit negative examples. The words highlighted in red are randomly replaced based on their phonetic similarity.}
    \label{fig_example2}
  
\end{figure}

\subsection{Data Assumption}
We assume availability of the following sources for training the entity-aware language model:

{\bf Raw text:} A large amount of raw text ${\cal X} = \{x^{(1)}, \ldots, x^{(m)}\}$ where $x^{(j)}$ is the $j$th sentence in the dataset. Each sentence $x^{(j)}$ consists of $l_j$ words: $\{x^{(j)}_1, \cdots, x^{(j)}_{l_j}\}$.  In this paper, we train our model on music queries. These queries are usually about asking a voice system to play or download some particular music.

{\bf Knowledge-base:} A database ${\cal K}=\{k^1, \cdots, k^n\}$ such that each entry $k^j$ in the knowledge-base consists of $m$ fields $k^j = \{k_1^j, \cdots, k_m^j\}$. In this paper, we use a knowledge-base that consists of music information. We use three fields: artist name, song title, and the frequency of usage of the song in our in-house application.

\subsection{Contrastive Estimation}
Contrastive estimation~\cite{smith2005contrastive} requires creation of \emph{implicit} negative examples for each training sample. This is done by injecting artificial noise to the correct example. Therefore, for every sentence $x^{(j)} \in {\cal X}$ in the training data, we approximate its probability with respect to the negative examples ${\cal N}(x^{(j)})$ and model parameter $\theta$. 

\begin{equation}\label{prob_eq}
\begin{split} 
P(x^{(j)} ; \theta) & \approx  P(x^{(j)} | {\cal N}(x^{(j)}); \theta) \\ 
& = \frac{e^{u(x^{(j)} ; \theta)}}{\sum\limits_{x' \in\{ {\cal N}(x^{(j)}) \cup x^{(j)}\} } e^{u(x' ; \theta)}} 
\end{split}
\end{equation}
where $u(x^{(j)} ; \theta)$ is the scoring function of the sentence $x^{(j)}$ given the model parameter $\theta$. The objective function for the training data ${\cal X}$ is the following:

\begin{equation}
L({\cal X} ; \theta) = - \sum_{j=1}^{m} \log  P(x^{(j)} ; \theta) + \lambda \norm{\theta}
\end{equation}
where $\lambda$ is a constant coefficient for L2 regularization.

\subsubsection{Creating Negative Examples}

The definition of the negative example function depends on the task. Since we target outputs from a voice system, our observation shows that most of the real errors come from a confusion by the model between two phonetically similar words or phrases. We use a simple phonetic similarity function to randomly pick words in a training sentence and replace them with one of their phonetically similar words. We rerank the negative samples based on their n-gram language model probability and pick the five highest scoring ones. This is mainly because the real n-best lists usually consist of relatively fluent sentences while so many of the negative samples are not fluent sentences. By applying this reranking step, we avoid totally irrelevant negative samples. Figure~\ref{fig_example2} shows a real example of the negative examples created by our method.

\subsubsection{Comparison to Noise Contrastive Estimation (NCE)} 
Noise contrastive estimation~\cite{pmlr-v9-gutmann10a} is a popular method in language modeling. In this method, the probability of a word is maximized given its negative samples. The negative samples come from a probability distribution aside from the current model. Then the method is defined as a binary classification problem in which the label for the correct word is one and for the negative examples is zero. NCE has interesting properties such as being self-normalized. One challenge in NCE is that one should be able to define a well-formed probability distribution for negative examples. This is very straightforward for word-level language modeling; for example, we can define the noise distribution as the categorical distribution derived from the word counts in the training data. In our case, we are interested in changing more than one word in a sentence to create negative examples. That makes it hard for us to define a well-formed probability distribution for negative examples. On the other hand, although contrastive estimation is not as principled as NCE, we do not have a strict limitation on defining the negative examples using contrastive estimation.

\subsection{Features}
For a sentence $x$ with $l$ words, the following features are used:

\begin{itemize}
    \item {\bf Recurrent representation:} We use a bidirectional LSTM (BiLSTM) $\beta$ to compute the sentence-level representation for each sentence. In other words, we have two independent LSTMs, one for the forward pass that sweeps a sentence from left to right and the other for the backward pass that does a reverse sweep of the sentence. The forward and backward LSTMs give the following outputs:
    \begin{equation}
    \begin{split}
          [E[x_1], \ldots, E[x_l]] \xrightarrow{f-LSTM} [h^f_1, \cdots, h^f_l] \\
       [E[x_l], \cdots, E[x_1]] \xrightarrow{b-LSTM} [h^b_l, \cdots, h^b_1] 
    \end{split}
    \end{equation}
    where $E[x_i]$ is the word embedding vector for word $x_i$.
 
    We use the concatenation of the final representations as the recurrent representation of the sentence. The LSTM parameters are updated during training with backpropagation:
    
    \begin{equation}
    \phi^{rnn} (x; \beta, E) = [h^f_l;  h^b_1]  
    \end{equation}

    \item {\bf N-gram LM probability:} This score is fixed during training:
    
    \begin{equation}
     \phi^{ngram} (x; \theta_{ngram}) = -\log P(x; \theta_{ngram})
     \end{equation}
     
     The ngram language model feature is obtained with 10-way jack-knifing in order to avoid overfitting to the training data.
    
     \item {\bf Phrase pair co-occurrence:} For every pair of non-overlapping phrases in sentence $x$, we count the number of entries in the knowledge-base that has the first subphrase as an \emph{artist} and the second as a \emph{song name} (or vice versa). Finally, we quantize this value into a bin of size $10$ (based on the maximum possible co-occurrence count in the knowledge-base) and embed that to a embedding dictionary $\gamma \in \mathbb{R}^{10 \times d_{m}}$. Thus the cross-entity phrase co-occurrence feature $\phi^{co} (x ; \gamma, {\cal K})$ is a $d_m$-dimensional vector.

      \item {\bf Subsentence knowledge-base frequency:} We compute the sum of \emph{frequency field} of each phrase in a sentence $x$ for the artist and song fields.
      \begin{equation}
      \begin{split}
          f^{1} (x; {\cal K}) &= \log  \sum^{l}_{i=1} \sum^{l}_{j=i} freq^{\cal K}_{artist} x[i:j] \\
          f^{2} (x; {\cal K}) &= \log  \sum^{l}_{i=1} \sum^{l}_{j=i} freq^{\cal K}_{song} x[i:j]
      \end{split}
	\end{equation}

	 We quantize these values into the integer range $[0,100)$ 	and represent them as embedding vectors $\mu, \nu \in \mathbb{R}^{100 \times d_{c_f}}$:
		\begin{equation} 
		\begin{split}
		    c_f^{1} (x; {\cal K}) &= \floor*{min (50\times f^{artist} (x), 99)} \\
		c_f^{2} (x; {\cal K}) &= \floor*{min (50\times f^{song} (x), 99)}
		\end{split}
		\end{equation}

	 The final features are $\phi^{1} (x; \mu, {\cal K})$ and $\phi^{2} (x; \nu, {\cal K})$ as two $d_f$ dimensional vectors.
    
    \item {\bf Cross-entity and Intra-entity word-based mutual information:} We observed that many phrasal entities have different forms (such as using abbreviations for first names). Therefore, it can be useful to incorporate word-level features for words in entities. This is done by enumerating the mutual information between words across the artist and song fields. We use the average of normalized pointwise mutual information~\cite{bouma2009normalized} between words. 
    \begin{equation}
    \begin{split}
         &   in (x; {\cal K}) =  \\
           &\frac{ \sum_{i=1}^{l-1}  \sum_{j=i+1}^{l} -\log \frac{p^{\cal K}(x_i,x_j)}{p^{\cal K}(x_i)p^{\cal K}(x_j)}/\log p^{\cal K}(x_i,x_j)}{l \cdot (l-1)/2}
    \end{split}
     \end{equation}
    
    where the probabilities $p^{\cal K}$ are calculated based on the frequency information in the knowledge-base. We scale the value $in(x,k)$ to be in $[0,1]$ and then quantize it into 100 bins:
    \begin{equation}
    f_{in}(x; {\cal K}) = \floor*{100 \cdot \frac{in (x) + 1}{2}} 
    \end{equation}
    
    Finally, we embed this to an embedding parameter $\eta \in \mathbb{R}^{100 \times d_{xp}}$. $\phi^{xp}(x; \eta, {\cal K}) \in \mathbb{R}^{d_{xp}}$ is the embedding feature for representing the cross-entity word-based mutual information. Similarly, we use an embedding dictionary $\kappa \in \mathbb{R}^{100 \times d_{ip}}$ and $\phi^{ip}(x; \kappa, {\cal K}) \in \mathbb{R}^{d_{ip}}$ for the feature representing intra-entity word-based mutual information.

\end{itemize}

% \begin{figure*}[t!]
% \floatbox[{\capbeside\thisfloatsetup{capbesideposition={left,top},capbesidewidth=4cm}}]{figure}[\FBwidth]
% {\caption{A graphical depiction of the scoring function for the reranker model. Here, we assume that the knowledge base and the ngram model $\theta_{ngram}$ is already given to us, and the model will learn the embedding and dense parameters using the negative examples during training. The parameters $E, \gamma, \nu, \mu, \eta, \kappa$ are embedding parameters, $\beta$ is the set of BiLSTM parameters; and $H$ and $\omega$ are dense layers. The scores from the ngram model is interpolated with the final score from the deep model. In training, there is a softmax layer on top that defines the contrastive estimation loss with respect to the correct output in contrast to the negative examples in the n-best list.}\label{fig_network}}
% {   \includegraphics[height=0.3\textheight,width=0.7\textwidth]{interspeech2018_model_struct}}
%  \end{figure*}

\begin{figure}[t!]
    \centering
    \includegraphics[height=0.3\textheight,width=1\textwidth]{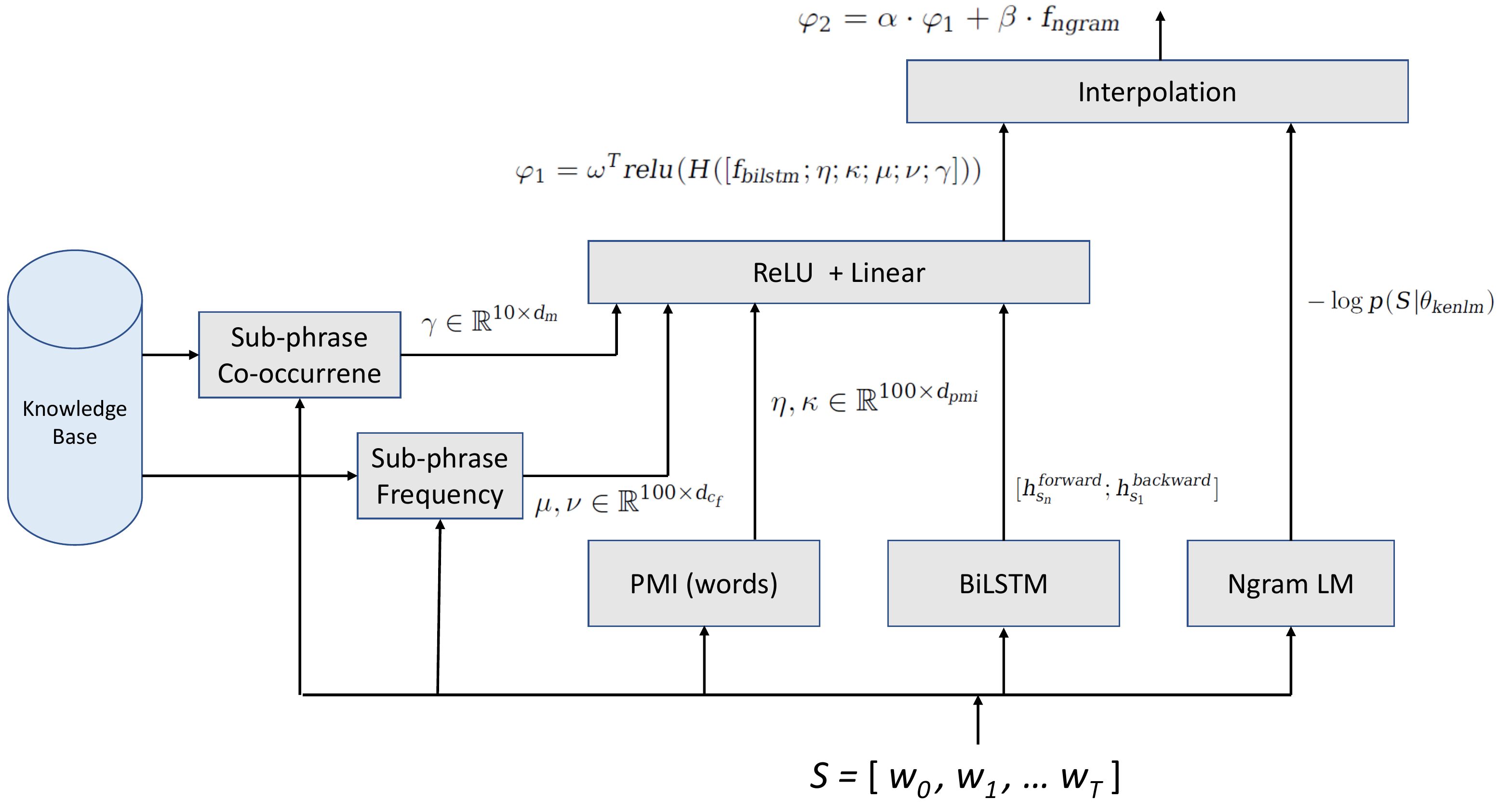}
    \caption{A graphical depiction of the scoring function for the reranker model. Here, we assume that the knowledge base and the ngram model $\theta_{ngram}$ is already given to us, and the model will learn the embedding and dense parameters using the negative examples during training. The parameters $E, \gamma, \nu, \mu, \eta, \kappa$ are embedding parameters, $\beta$ is the set of BiLSTM parameters; and $H$ and $\omega$ are dense layers. The scores from the ngram model is interpolated with the final score from the deep model. In training, there is a softmax layer on top that defines the contrastive estimation loss with respect to the correct output in contrast to the negative examples in the n-best list.}
    \label{fig_network}
\end{figure}
% \begin{figure}[t!]
% \includegraphics[height=0.5\textheight,width=1.1\textwidth]{network}

% \caption{A graphical depiction of the scoring function for the reranker model. Here, we assume that the knowledge base ${\cal K}$ and the ngram model $\theta_{ngram}$ is already given to us; and the model will learn the embedding and dense parameters using the negative examples during training. The parameters $E, \gamma, \nu, \mu, \eta, \kappa$ are embedding parameters, $\beta$ is the set of BiLSTM parameters; and $H$ and $\omega$ are dense layers. The scores from the ngram model $\phi^{ngram}(x; \theta_{ngram})$ is interpolated with the final score from the deep model. In training, there is a softmax layer on top that defines the contrastive estimation loss with respect to the correct output in contrast to the negative examples in the n-best list.}\label{fig_network}
% \end{figure}

\subsection{Network Architecture}
All the features, except the ngram probability, are concatenated as $\phi(x ; \theta, {\cal K})$ and fed to a hidden layer with a rectified linear unit (RELU)~\cite{nair2010rectified} activation. The output from the hidden layer is multiplied by a vector $\omega$:

\begin{equation}
s (x ; \theta,{\cal K}) = \omega^{T} relu ( H^{T} \phi(x ; \theta, {\cal K}) )
\end{equation}

Finally, we use a linear interpolation between $h(x, \theta)$ and the n-gram feature $\phi^{ngram}(x; \theta)$. All parameters in this model, except the ngram probabilities, are tuned during backpropagation:
\begin{equation}\label{u_eq}
u(x ; \theta, {\cal K}, \theta_{ngram}) = \alpha_1 \cdot s(x ; \theta, {\cal K}) + \alpha_2 \cdot \phi^{ngram}(x; \theta_{ngram})
\end{equation}
where during training, we use the parameter $u$ in the above equation (Eq.~\ref{u_eq}) to calculate the approximate probability of $x$ with respect to negative examples as in Eq.~\ref{prob_eq}.  

Figure~\ref{fig_network} shows a graphical depiction of the network structure.

%In summary, the model parameters are: 
%\[
%\theta = \{\alpha_1,\alpha_2, \beta, E, \theta_{ngram}, \gamma, \nu, \mu, \eta, \kappa\}
%\]

\section{Experiments}\label{sec_experiments}
%This section describes the experimental settings and results. 

\subsection{Data and Setting}
Our pipeline uses a domain classifier to classify queries in a query stream. We select queries tagged as belonging to the music domain. We mapped words with frequency less than one in the training data to the \emph{unknown} symbol. Our heldout data is a small set of transcribed queries ($3941$ sentences) from a mobile phone application and our blind test set is from a distant microphone with relatively poor quality speech signals ($4970$ sentences). Both the heldout and test data have five hypotheses per sentence.  The knowledge-base consists of 13 million entries. For each entry, we select the title of the song, artist name,  and the frequency of requests. The training data has $5.2$ million sentences consisting of 24 million words. The vocabulary size after converting infrequent words to the unknown symbol is $129935$. %The heldout data consists of $3941$ sentences. The blind test set contain $4970$ sentences. 

\subsection{Negative examples}
We first train the KenLM ngram language model \cite{kenlm} with 10-way jackknifing on the training data.  We sample 30 random negative examples for each sentence in the training data and use the probabilities from the language model to get the 5 best examples. Our observation shows that the quality of some of the negative examples are still not promising. Therefore, we just kept the top one million training examples where the average ngram probability of their negative examples are highest. We train the same ngram model on the training data to calculate probabilities on the heldout and test data.  

\subsection{LSTM language model and reranker}
Since our reranker just uses a portion of the training data, we might lose some performance from having a smaller training data. Therefore, the final score of each sentence in decoding is summed with the score from an LSTM model trained on the whole training data. To train the standard language model with LSTM, we use noise contrastive estimation~\cite{pmlr-v9-gutmann10a} with $100$ fixed samples per each minibatch of $64$ sentences to train our model. The bias term is initialized as $-\log N$ where $N$ is the vocabulary size. We use L2-regularization with coefficient $10^{-6}$. Stochastic gradient descent with momentum~\cite{Hinton2012} and learning rate of $1.0$ with momentum $0.9$ and decay $0.5$ is used to train the model parameters. We apply dropout~\cite{srivastava2014dropout} with probability $0.2$ in training. Word embedding vectors are initialized randomly with dimension $200$ and the LSTM dimension is $1000$.

We use a similar LSTM model in both directions inside the reranker but with a smaller dimension for the LSTM representation ($500$). Early stopping is applied based on word error rate improvement on the heldout data. We use dimension of $50$ for $d_f$, $d_{xp}$ and $d_{ip}$ and $10$ for $d_m$.  We use the Dynet library~\cite{neubig2017dynet} for implementing all of our models.

% \begin{table}[]
%     \centering
%     \begin{tabular}{l|c}
%          \hline \hline
%         Ngram model & KenLM~\cite{kenlm} \\
%         Trainer & Momentum SGD  \\
%         Momentum & $0.9$ \\
%         Learning rate & $1.0$ \\
%         Decay & $0.5$\\ 
%         $\lambda$ (L2 regularization) & $10^{-6}$\\
%         Dropout & $0.2$\\
%         Word embedding & $200$ \\
%         LSTM dimension & $500$ \\
%       $d_m$ & $10$ \\
%       $d_f$ & $50$ \\
%       $d_{xp}$ & $50$ \\
%       $d_{ip}$ & $50$ \\
%             \hline \hline
%     \end{tabular}
%     \caption{Parameters used in the reranker model.}
%     \label{tab_params}
% \end{table}

\subsection{Results}
Table~\ref{tab_results} shows the experimental results on the heldout and test data. As shown in the table, the reranker, with only one-fifth of the real training data, perform better than both the LSTM and ngram language models.  When the reranker is merged with the LSTM language model, the performance improves further. We believe that is due to the bigger size of the training data for the LSTM model. The final ensemble result achieves an absolute $0.44\%$ improvement compared to the LSTM language model on the test data. Another observation is that LSTM LM is more effective than ngram LM for reranking on the heldout set but not on the test set (rows 2 and 3 in Table~\ref{tab_results}). A possible explanation of this result is the difference in quality of the n-best hypotheses between heldout and test sets. Recall that the speech for heldout set is collected from mobile phones whereas the test set data is collected from a distant device. 

\begin{table}[t!]
    \centering
    \begin{tabular}{l | c | c}
    \hline \hline
        Model & heldout & Test \\ \hline
        First-pass (no reranker) & 11.52 & 12.52 \\
        Ngram LM & 11.17 & 11.56 \\
        LSTM LM & 10.16 & 11.56 \\
        Reranker & 10.03 & 11.35 \\
        Reranker + LSTM & {\bf 9.82} & {\bf 11.12} \\ \hline
        Oracle & 6.55 & 7.32 \\
        \hline \hline
    \end{tabular}
    \caption{Experimental results based on word error rate (WER) on the heldout and test data. First-pass shows the model accuracy by just applying the first-pass language model from the speech recognition system. The other rows show the effect of interpolating it with different types of language models or reranker. Oracle shows the WER when we pick the sentence in the n-best list with the lowest WER with respect to the gold-standard output.}
    \label{tab_results}
\end{table}
\section{Conclusion}\label{sec_conclusion}
In this paper, we have shown an effective method to encode real-world entities into a language model.  We designed a simple but effective approach to create artificial n-best lists, thus obviating the need for annotated data. Our experiments show an improvement on the heldout and test datasets.  One interesting direction to pursue is to incorporate a small amount of transcribed data and use our approach on a combined set of transcribed dataset and artificial training data. There are certain challenges facing this approach when dealing with a mixture of real n-best lists and artificial n-best lists such as deciding about the proportion of real n-best lists compared to the artificial ones. Future work should consider studying this problem.

\section{Acknowledgement}
We thank the anonymous reviewers for their valuable feedback. This research was conducted while the first author was an intern in the language modeling group at Microsoft in Sunnyvale, California. We would like to thank the researchers in the group for helpful discussions and assistance on different aspects of the problem.

\bibliography{refs}
\bibliographystyle{IEEEtran}

\end{document}